%% file: main.tex
\documentclass[letterpaper,10pt,conference]{ieeeconf}  % Comment this line out if you need a4paper

\IEEEoverridecommandlockouts                              % This command is only needed if 
\usepackage{mymacros}
\usepackage{times} % assumes new font selection scheme installed
\usepackage{graphicx}

\usepackage[T1]{fontenc}
\usepackage{color}
\usepackage{hyperref}
\hypersetup{
 setpagesize=false,
 bookmarksnumbered=true,%
 bookmarksopen=true,%
 colorlinks=true,%
 linkcolor=blue,
 citecolor=red,
}
\usepackage{pifont}% http://ctan.org/pkg/pifont
\usepackage{subcaption}

\let\labelindent\relax
\usepackage{enumitem}

\title{\LARGE \bf
Learning-To-Rank Approach for Identifying Everyday Objects \\Using a Physical-World Search Engine
}

\author{Kanta Kaneda, Shunya Nagashima, Ryosuke Korekata, Motonari Kambara and Komei Sugiura% <-this % stops a space
\thanks{Authors are with Keio University, 3-14-1 Hiyoshi, Kohoku, Yokohama, Kanagawa 223-8522, Japan.
         {\tt\small \{k.kaneda, ng\_sh, rkorekata, motonari.k714, komei.sugiura\}@keio.jp}
}
}

\begin{document}

\maketitle
\thispagestyle{empty}
\pagestyle{empty}

\input{abstract}
\input{section1}
\input{section2}

\input{section3}
\input{section4}

\input{section5}

\input{section6}

\input{section6-2}
\input{section7}

% \appendix
% \input{sectionA}

\section*{ACKNOWLEDGMENT}
This work was partially supported by JSPS KAKENHI Grant Number 20H04269, JST Moonshot, and NEDO.

\bibliographystyle{IEEEtran}
\bibliography{reference}

\end{document}

%% file: abstract.tex
\begin{abstract}
Domestic service robots offer a solution to the increasing demand for daily care and support. A human-in-the-loop approach that combines automation and operator intervention is considered to be a realistic approach to their use in society. 
Therefore, we focus on the task of retrieving target objects from open-vocabulary user instructions in a human-in-the-loop setting, which we define as the learning-to-rank physical objects (LTRPO) task. For example, given the instruction ``Please go to the dining room which has a round table. Pick up the bottle on it,’’  the model is required to output a ranked list of target objects that the operator/user can select.
In this paper, we propose MultiRankIt, which is a novel approach for the LTRPO task.
MultiRankIt introduces the Crossmodal Noun Phrase Encoder to model the relationship between phrases that contain referring expressions and the target bounding box, and the Crossmodal Region Feature Encoder to model the relationship between the target object and multiple images of its surrounding contextual environment. 
Additionally, we built a new dataset for the LTRPO task that consists of instructions with complex referring expressions accompanied by real indoor environmental images that feature various target objects.
We validated our model on the dataset and it outperformed the baseline method in terms of the mean reciprocal rank and recall@k.
Furthermore, we conducted physical experiments in a setting where a domestic service robot retrieved everyday objects in a standardized domestic environment, based on users' instruction in a human--in--the--loop setting. The experimental results demonstrate that the success rate for object retrieval achieved 80\%.
Our code is available at this URL\footnote{https://github.com/keio-smilab23/MultiRankIt}.
\end{abstract}

%% file: section1.tex
\section{Introduction
\label{intro}
}
\vspace{-1mm}

The aging of the population has led to an increased demand for daily care and support. The shortage of home care workers has become a pressing societal problem, and the use of domestic service robots (DSRs) to assist individuals with disabilities is considered to be a possible solution.
A realistic approach to using DSRs for societal applications is a human-in-the-loop setting that combines automation and intervention by their operator.
Therefore, in this study, we consider that a DSR that is neither fully automated nor fully teleoperated, but rather serves as a cybernetic avatar\cite{ishiguro2021realisation} for the user.
In this setting, a DSR that could provide relevant options to the operator would be highly beneficial.

In this study, we focus on the task of retrieving target objects from open-vocabulary user instructions, for example, manipulating or moving objects in the domestic environment, in a human-in-the-loop setting.
We define this task as a learning-to-rank physical objects (LTRPO) task.
Fig.~\ref{fig:overview} shows a typical scene for this task. Given the instruction ``Go to the bathroom with a picture of a wagon. Bring me the towel under the picture directly across from the sink,'' the model is required to output a ranked list of target objects.
In the human-in-the-loop setting, it is important to display and rank a number of target objects that are easily selectable by the user/operator because human attention is a limited resource. Additionally, displaying a limited number of objects helps to minimize the cognitive load on the user/operator.

\input{fig/overview}

The LTRPO task is challenging because of its difficulty in accurately retrieving the specified target object from complex instructions.
Specifically, the success rate for humans in a referring expression comprehension (REC) task with a simple referring expression is 90.76\%, whereas it is only 48.98\% for a state-of-the-art model \cite{yu2018mattnet}.
% However, the LTRPO task is more challenging than the REC task because it involves identifying the target object from instructions with multiple referring expressions (e.g., ``Please go to the dining room which has a round table. Pick up the bottle on it'').

The LTRPO task differs from REC tasks in that its objective is to retrieve images of the target object from complex instructions, often spanning more than two sentences and incorporating multiple referring expressions (see Fig.\ref{fig:overview}).
% (e.g., ``Go to the bathroom with a picture of a wagon. Bring me the towel under the picture directly across from the sink'').
Note that the average expression length of a common REC task dataset is 8.43 words \cite{Mao_2016_CVPR}, while the average sentence length of the LTRPO dataset is 18.8 words, highlighting the significantly more complex nature of sentences in the LTRPO task.
Additionally, while the REC task focuses on localizing the target object from a single image, the LTRPO task aims to identify the target object from hundreds of images within the environment. The human--in--the--loop setting changes the goal of the solution, allowing the user/operator to select the desired target object from the output ranked image list.
 Consequently, the methods proposed for the REC task are impractical for the LTRPO task.

Recent advancements in multimodal representation learning, such as CLIP\cite{radford2021learning}, have resulted in the improved performance of crossmodal retrieval. However, as explained above, the performance of the crossmodal retrieval using texts that contain complex referring expressions remains insufficient. This is because existing methods often consider tasks that use only simple expressions (e.g., ``A photo of \{label\}'' \cite{radford2021learning}).

In this paper, we propose MultiRankIt, a method to retrieve target objects from open-vocabulary user instructions in a human-in-the-loop setting. 
% Our code and a demonstration video are available at these URLs\footnote{Available upon acceptance}$^,$\footnote{Available upon acceptance}.
In MultiRankIt, we introduce the Crossmodal Noun Phrase Encoder (CNPE), which handles noun phrases and prepositional phrases extracted from the instruction to model the relationship between the extracted phrases and the target bounding box.
By introducing this module, it is expected that the model can retrieve the appropriate target object from an instruction with complex referring expressions.
We also introduce the Crossmodal Region Feature Encoder (CRFE), which handles multiple images of the surrounding context.
By handling context images that capture a broader range, rather than only those that are in close proximity to the target bounding box, it is expected that the model can effectively model the relationship between the target object and its surrounding contextual environment.

The difference between our method and existing methods is that our method considers the task as a learning-to-rank task \cite{liu2009learning} in which a DSR generates a ranked list of candidate target objects that are presented to the user. 
The proposed approach aims to strike a balance between automation and human intervention, where not everything is performed automatically and conversely, not everything is performed manually. 

Our main contributions  are summarized as follows:
% \begin{itemize}
%  \item We propose MultiRankIt, which is a novel approach for identifying target objects from open-vocabulary user instructions in a human-in-the-loop setting.
%  \item We introduce the CNPE to model the relationship between phrases that contain referring expressions and target bounding box.
%  \item We introduce the CRFE to model the relationship between the target object and multiple images of its surrounding contextual environment.
% \end{itemize}

\begin{itemize}
 \item We propose MultiRankIt, which is a novel approach for identifying target objects from user instructions in a human-in-the-loop setting. While existing methods that address object localization as an image retrieval task (e.g., \cite{hu2016natural}) are limited to a closed-vocabulary setting,  MultiRankIt handles open-vocabulary user instructions. In addition, to the best of our knowledge, this is the first attempt to propose this setting for mobile manipulation.
 \item We introduce the CNPE to model the relationship between phrases that contain referring expressions and target bounding boxes. While existing methods (e.g.,  \cite{subramanian2022reclip,baldrati2023multimodal,cartella2023openfashionclip}) are limited to handling simple noun phrases referring solely to the target object, the proposed method can accommodate various and hierarchical referring expressions within the instruction. This allows for the effective modeling of referring expressions related to both the target object and its surrounding environment.
 \item We introduce the CRFE to model the relationship between the target object and multiple images of its surrounding contextual environment. While existing methods (e.g., \cite{hatori2018interactively,ishikawa2021target}) are unable to consider relationships when dealing with referring expressions related to objects beyond the image frame, the proposed method can adequately model the relationships with surrounding objects and landmarks across a broader viewpoint.

\end{itemize}

%% file: fig/overview.tex
\begin{figure}[t]
    \vspace{2mm}
    \centering
    \includegraphics[width=83mm]{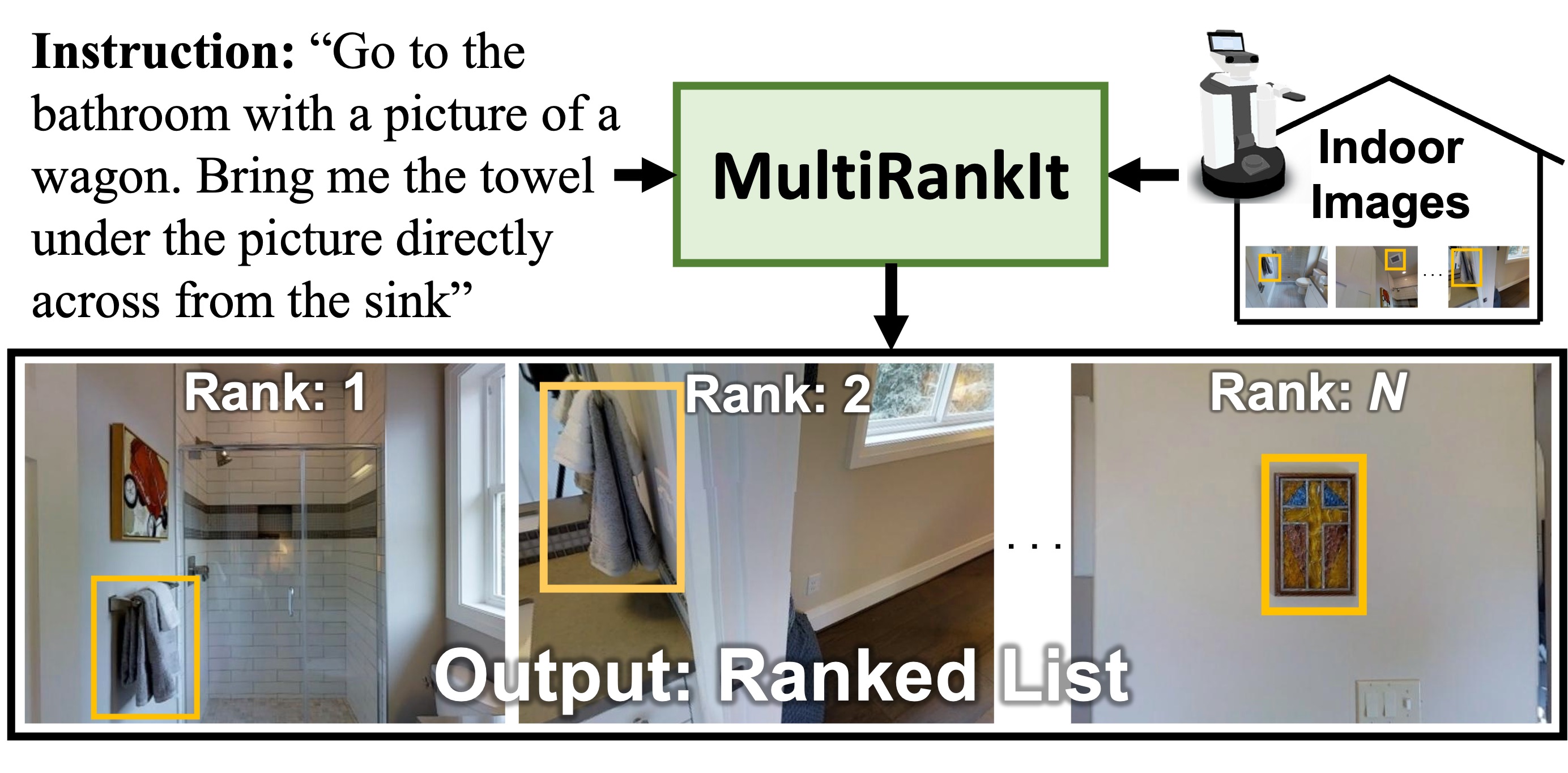}
    \vspace{-2mm}
    \caption{Typical scene of the LTRPO task. The input is an instruction and output is the ranked list of target objects.}
    \label{fig:overview}
    \vspace{-6mm}
\end{figure}

% containing complex referring expression
% such as ``Go to the bathroom with a picture of a wagon. Bring me the towel under the picture directly across from the sink,''

%% file: section2.tex
\vspace{-2.5mm}
\section{Related Work
\label{related}
}
\vspace{-1mm}

Numerous studies have been conducted in the field of vision and language \cite{uppal2022multimodal,ishikawa2021target,yu2019deep,li2019relation,yang2020bert,dlct,m2trm,anderson2021sim,wang2022counterfactual,chen2022reinforced,weyand2020google,vo2019composing,kim2021dual,shridhar2022cliport,zhou2023modularity,magassouba2019understanding,ishikawa2022moment,hatori2018interactively,shah2023lm,huang2023visual}.
In \cite{uppal2022multimodal}, the authors provide a comprehensive overview of various tasks, and the latest methods for vision and language tasks.
% Extensive studies in the field of vision and language have been conducted in recent years, and encompass tasks such as visual question answering \cite{zhou2020unified,yu2019deep,li2019relation}, image captioning \cite{yang2020bert,dlct,m2trm}, and vision language navigation\cite{anderson2021sim,wang2022counterfactual,chen2022reinforced}.
% In the following, we discuss related research in the field of vision and language, and focus on vision and language pretraining (VLP) \cite{radford2021learning}, crossmodal retrieval \cite{weyand2020google,wu2021fashion,vo2019composing,kim2021dual,baldrati2022conditioned} and vision and language for robotics field \cite{ishikawa2021target,magassouba2019understanding,ishikawa2022moment,hatori2018interactively}.
% In the following, we discuss related research in the field of vision and language, and focus on  crossmodal retrieval \cite{weyand2020google,wu2021fashion,vo2019composing,kim2021dual,baldrati2022conditioned} and vision and language for robotics field \cite{shridhar2022cliport,zhou2023modularity,magassouba2019understanding,ishikawa2022moment,hatori2018interactively}.
In the following, we discuss related research in the field of vision and language, and focus on  crossmodal retrieval, vision and language for robotics field and REC.

% The field of VLP aims to explore the semantic relationship between visual and textual modalities through pretraining using large-scale datasets. CLIP \cite{radford2021learning} is a typical method in VLP, that adopts a contrastive approach to learn image representations from text by maximizing the similarity between correct text-image pair embeddings. It has been trained on a corpus of 400M image-text pairs and demonstrated strong zero-shot transfer abilities in various tasks, including composed image retrieval \cite{baldrati2022conditioned}, image generation \cite{ramesh2022hierarchical} and embodied AI \cite{khandelwal2022simple}.

Crossmodal retrieval involves the retrieval of samples from one modality based on a query expressed in another modality. 
% This task has a wide range of applications, including retrieval of landmarks \cite{weyand2020google} and fashion items\cite{wu2021fashion}.
In \cite{vo2019composing}, the authors present an image retrieval method that uses an input query in the form of an image and an accompanying text description of desired modifications. They introduce Text Image Residual Gating (TIRG), which combines the reference image and text query using a gating function and residual connection.
DCNet \cite{kim2021dual} extends methods such as TIRG by introducing the Correction Network, which incorporates the difference between the target
and reference image to learn robust multimodal representations.
% CLIP4CIR \cite{baldrati2022conditioned} extends the CLIP method for crossmodal retrieval tasks. The model uses a pretrained CLIP image encoder and a fine-tuned CLIP text encoder. 
% It achieved state-of-the-art performance on the CIRR \cite{liu2021image} and FashionIQ \cite{wu2021fashion} datasets.

Several standard datasets exist that handle the crossmodal retrieval task (e.g., \cite{han2017automatic,liu2021image}).
The CIRR dataset \cite{liu2021image} consists of over 36,000 pairs of open-domain images paired with human-generated modifying text. It comprises real-life images that are not restricted to a specific domain, and places greater emphasis on distinguishing between visually similar images.
The Fashion200k dataset \cite{han2017automatic} consists of over 200,000 images of fashion products. The dataset provides human-generated captions that distinguish similar pairs of fashion images, along with side information that consists of real-world product descriptions.

\input{fig/model.tex}

% Additionally, numerous studies have explored visual-language models in robotic applications, including
Additionally, studies related to vision and language for the robotics field include navigation \cite{shah2023lm,huang2023visual} and object manipulation \cite{shridhar2022cliport,zhou2023modularity,magassouba2019understanding,ishikawa2022moment}.
LM-Nav \cite{shah2023lm} introduces a robotic navigation system that effectively utilizes ViNG\cite{shah2021ving}, CLIP\cite{radford2021learning} and GPT-3\cite{brown2020language} to achieve successful long-horizon navigation in real outdoor environments based on natural language instructions.
CLIPort \cite{shridhar2022cliport} combines semantic understanding from CLIP with spatial precision from Transporter\cite{zeng2021transporter}, allowing successful execution of tabletop tasks without explicit representations of object poses, memory, or syntactic structures.
% Additionally, studies related to vision and language for the robotics field include the following:
In \cite{magassouba2019understanding}, authors address the multimodal language understanding task for fetching instructions, in which the goal is to predict the target object as instructed by the user from an unconstrained sentence. 
% Target-dependent UNITER \cite{ishikawa2021target} introduces a UNITER-based transformer architecture to model the relationship between text and visual features.
HLSM-MAT \cite{ishikawa2022moment} addresses the vision and language task in the execution of household tasks based on textual instructions using the  Moment-based Adversarial Training algorithm, which uses two types of moments for perturbation updates.

The REC task shares some similarities with the LTRPO task.
The REC task requires an agent to localize an object in an image given a natural language expression. RefTeacher \cite{sun2023refteacher} employs a semi-supervised teacher-student learning paradigm with Attention-based Imitation Learning and Adaptive Pseudo-label Weighting, achieving near fully supervised results using only 10\% labeled data.
% to substantially improve REC performance, achieving near fully supervised results using only 10\% labeled data.
Target-Dependent UNITER \cite{ishikawa2021target} uses the transformer attention mechanism based on the UNITER \cite{chen2020uniter} framework to model the relationship between objects and instructions for multimodal understanding for fetching instructions.
However, as explained in Section \ref{intro}, methods proposed for the REC task are not directly applicable to the LTRPO task.

As mentioned above, our method differs from existing methods (e.g.,  \cite{vo2019composing,kim2021dual}) in that we consider the task as a learning-to-rank task \cite{liu2009learning} in which a DSR generates a ranked list of candidate target objects that are presented to the user/operator. 
 In addition, while existing methods typically involve a robot operating in an environment where it has no prior information about its surroundings, the LTRPO task is specifically designed for situations where the robot already has access to images of the environment. This particular setup is highly practical for domestic service robots that are designed to operate within the same environment over extended periods. 
Our method is also different from \cite{hatori2018interactively} in that it retrieves the target object from multiple images showing the domestic environment, whereas in \cite{hatori2018interactively}, the method identifies the target object from a single image with a fixed viewpoint. Additionally, our method introduces a module to model the relationship between the target object and multiple images of its surrounding contextual environment.
Furthermore, our method differs from existing methods that handle noun phrases (e.g., \cite{subramanian2022reclip,baldrati2023multimodal,cartella2023openfashionclip}) by introducing a module to model the relationship between phrases that contain referring expressions and target bounding box.

%% file: fig/model.tex
\begin{figure*}[t]
    \centering
    \includegraphics[width=160mm]{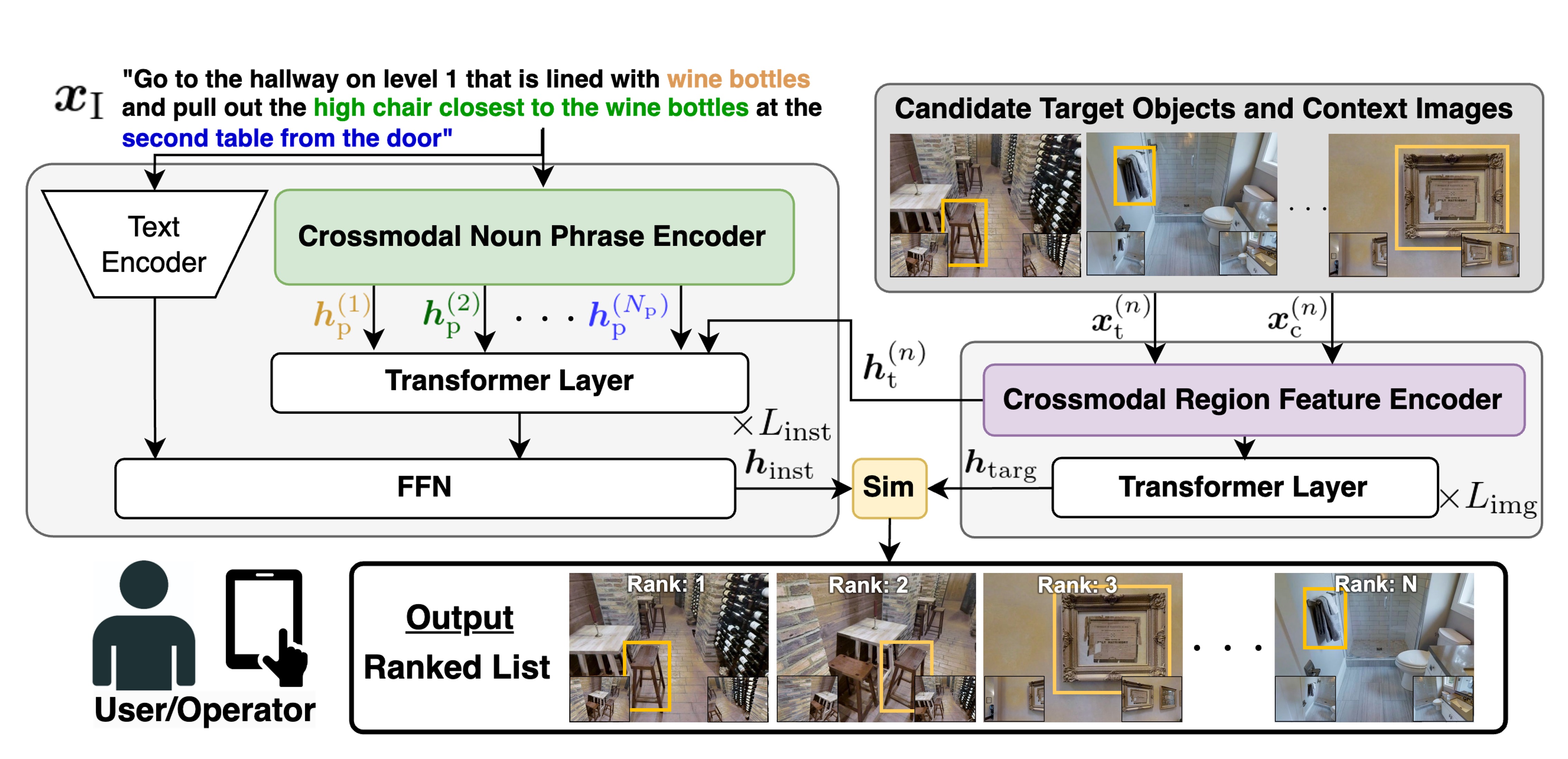}
    \vspace{-3mm}
    \caption{Proposed method framework. MultiRankIt consists of Crossmodal Noun Phrase Encoder and Crossmodal Region Feature Enconder. ``FFN'' and ``Sim'' denote the feedforward network and cosine similarity, respectively.}
    \label{fig:model}
    \vspace{-7mm}
\end{figure*}

%% file: section3.tex
\vspace{-1mm}
\section{Problem Statement
\label{sec:problem}
}
\vspace{-1mm}

% In this paper, we define the task as learning-to-rank physical objects (LTRPO) task. In this task, it is required to retrieve images of the target object taken by domestic service robots, according to given instructions.
% In this paper, we define learning-to-rank physical objects (LTRPO) task as the task to retrieve images of the target object taken by domestic service robots, according to given instructions.
In this paper, we define the LTRPO task as the task of retrieving images of the target object taken by DSRs according to open-vocabulary user instructions.
% In this task, it is desirable that appropriate target images are ranked high in the output rank list.
% The terminology used in this paper is defined as follows:
% Instruction I: A statement to domestic service robots to perform household tasks.
% Target object: The object that is the target of the instruction.
% Target bounding box: A bounding box of the target object. Note that there are N target bounding boxes T = {t₁, …, t_N} to be ranked by relevance.
% Context images: Images of the target object and its surroundings. Note that there are N context images C = {x_c₁, …, c_N}
The terminology used in this paper is defined as follows:
\begin{itemize}
    % \vspace{-1mm}
    \item \textbf{Instruction ${\bm x}_{\rm I}$:} an open--vocabulary instruction to DSRs to perform household tasks.
    \item \textbf{Target object:} the object that is the target of the instruction.
    
    \item \textbf{Target bounding box:} the bounding box of the target object. Note that there are $N_{\rm targ}$ bounding boxes in an environment to be ranked by relevance.  ($T = \{{\bm x}_{\rm t}^{(n)} | n = 1,2, ..., N_{\rm targ}\}$)
    \item \textbf{Context images:} images of the target object and its surroundings. For each ${\bm x}_{\rm t}^{(n)}$, there exists a corresponding ${\bm x}_{\rm c}^{(n)}$, resulting in a total of $N_{\rm targ}$ surrounding images in the environment. Note that duplication of images might be included in this set. ($C = \{ {\bm x}_{\rm c}^{(n)} | n = 1,2,..., N_{\rm targ}\}$) 
\end{itemize}
% \vspace{-4mm}

% Fig.~\ref{fig:overview} shows a typical scene for this task. Given the instruction ``Go to the bathroom with a picture of a wagon. Bring me the towel under the picture directly across from the sink,'' the model is required to output a ranked list of $T$.	
% The input and output of this task are defined as follows:
% % Input: Elements xᵢ = (I, tᵢ, ci) (i=1,2,...,N)
% % Output: Ranked list of T.
% \begin{itemize}
%     \item \textbf{Input:} Elements ${\bm x} = ({\bm x}_{\rm I}, T, C)$
%     \item \textbf{Output:} Ranked list of $T$.
% \end{itemize}
The input and output of this task are elements ${\bm x} = ({\bm x}_{\rm I}, T, C)$ and ranked list of $T$, respectively.
Note that the details for the input and output are explained in Section \ref{method}.
We assume that the coordinates of the target objects in the images are given.

%% file: section4.tex
\vspace{-2mm}
\section{Proposed Method
\label{method}
}
\vspace{-1mm}

% \subsection{Novelty}

In this paper, we propose MultiRankIt, which is a novel approach for identifying target objects from open-vocabulary user instructions, for example, manipulating or moving objects in the domestic environment, in a human-in-the-loop setting.
% We consider this as a learning-to-rank task \cite{liu2009learning} where a DSR generates a ranked list of candidate target objects to be presented to the user/operator. The user/operator selects the intended object from the list.
The novelty of this study, that is, the introduction of a learning-to-rank approach to multimodal object retrieval in a human-in-the-loop setting, could be applied to other tasks, such as multimodal image retrieval (e.g., \cite{wu2021fashion}).

% The novelty of our method are as follows:
% \begin{itemize}
%     \item We handle object search with human-in-the-loop settings in the home environment as an LTRPO task.
%     \item We introduce the CNPE which handles noun phrases with prepositional phrases. Unlike existing methods (e.g. \cite{subramanian2022reclip}), our method models the relationship between noun phrases and target bounding boxes.
%     \item We introduce the CRFE to model the relationship between the target object and multiple images of its surrounding contextual environment.
% \end{itemize}
% TODO

Fig.~\ref{fig:model} shows the structure of our method.
The proposed method consists of two main modules: CNPE and CRFE. The details of the modules are explained below.

% \input{fig/model.tex}

% \subsection{MultiRankIt}

Input $\bm{x}$ to our method is defined as follows:
\vspace{-2mm}
\begin{align*}
    {\bm x} &= ({\bm x}_{\rm I}, T, C), \\
    T &= \{{\bm x}_{\rm t}^{(n)} | n = 1,2, ..., N_{\rm targ}\}, \\
    C &= \{{\bm x}_{\rm c}^{(n)} | n = 1,2, ..., N_{\rm targ}\},
\end{align*}
where ${\bm x}_{\rm I} \in {\mathbb R}^{V \times L}$, ${\bm x}_{\rm t}^{(n)} \in {\mathbb R}^{3 \times W \times H}$ and ${\bm x}_{\rm c}^{(n)} \in {\mathbb R}^{N_c \times 3 \times 256 \times 256}$ denote an instruction tokenized as a one-hot vector with vocabulary size $V$ and maximum token length $L$, the bounding box of the target object of width $W$ and height $H$, and an image obtained by concatenating $N_c$ surrounding images of ${\bm x}_{\rm t}^{(n)}$, respectively.

% where ${\bm x}_{\rm I} \in {\mathbb R}^{V \times L}$ and ${\bm x}_{\rm t}^{(n)} \in {\mathbb R}^{3 \times W \times H}$ denote an instruction tokenized as a one-hot vector with vocabulary size $V$ and maximum token length $L$, and the bounding box of the target object of width $W$ and height $H$, respectively. 
% ${\bm x}_{\rm c}^{(n)} \in {\mathbb R}^{3 \times 256 \times 256}$,  ${\bm x}_{\rm l}^{(n)} \in {\mathbb R}^{3 \times 256 \times 256}$, and ${\bm x}_{\rm r}^{(n)} \in {\mathbb R}^{3 \times 256 \times 256}$ denote an image that contains the target bounding box, an image to the left of ${\bm x}_{\rm c}^{(n)}$, and an image to the right of ${\bm x}_{\rm c}^{(n)}$, respectively.
% Note that there are $N_{\rm targ}$ target bounding boxes and context images to be ranked.

% \subsection{MultiRankIt}
% \vspace{-1mm}

% Our model takes ${\bm x}_{\rm I}$, $T$ and $C$ as input, and outputs a ranked list of $T$. 
First, we construct a query $( {\bm x}_{\rm I}, {\bm x}_{\rm t}^{(n)}, {\bm x}_{\rm c}^{(n)} )$ to compute the similarity between ${\bm x}_{\rm I}$ and $({\bm x}_{\rm t}^{(n)}, {\bm x}_{\rm c}^{(n)})$.
As mentioned in Section \ref{intro}, many of the instructions in the LTRPO task contain complex referring expressions related to, for example, target objects, landmarks, and rooms. 
Therefore, we introduce the CNPE, which handles noun phrases and prepositional phrases extracted from ${\bm x}_{\rm I}$ to model the relationship between the extracted phrases and ${\bm x}_{\rm t}$.

The CNPE first extracts $N_p$ phrases from ${\bm x}_{\rm I}$ through the following steps:
First, it extracts noun phrases and prepositional phrases using the Stanford Parser \cite{schuster2016enhanced}. Then, we obtain $N_{\rm p}$ phrases ${\bm x}_{\rm p}^{(k)} (k=1,2,...,N_{\rm p})$ by grouping adjacent noun phrases and prepositional phrases.
We then obtain text features ${\bm h}_{\rm I} \in {\mathbb R}^{768}$ and ${\bm h}_{\rm p}^{(k)} \in {\mathbb R}^{768}$ from ${\bm x}_{\rm I}$ and ${\bm x}_{\rm p}^{(k)}$, respectively, using a pretrained CLIP text encoder \cite{radford2021learning}.
% Next, we model the relationship between ${\bm h}_{\rm p}^{(k)}$ and ${\bm h}_{\rm t}^{(n)}$ as follows:
% \begin{align}
%     {\bm h}_{\rm p} &= f_{\rm inst} ( {\bm h}_{\rm p}^{(1)}; {\bm h}_{\rm p}^{(2)}; ... {\bm h}_{\rm p}^{(k)}; {\bm h}_{\rm t}^{(n)} ), 
% \end{align}
% where $f_{\rm inst}$ and ${\bm h}_{\rm t}^{(n)}$ denote $L_{\rm inst}$ transformer layers and visual feature extracted from ${\bm x}_{\rm t}^{(n)}$ (details are explained later), respectively. Note that each  transformer layer consists of multi-head self attention layers \cite{vaswani2017attention} and feedforward networks.

Finally, we obtain the embedding for instruction ${\bm h}_{\rm inst}$ as follows:
\vspace{-2mm}
\begin{align*}
{\bm h}_{\rm inst} &= f_{\rm FFN} \left(
    \left[ f_{\rm trm} \left(
        \left[
        {\bm h}_{\rm p}^{(1)};  ... {\bm h}_{\rm p}^{(N_{\rm p})} \right] \right) ;{\bm h}_{\rm I};  {\bm h}_{\rm t}^{(n)}
    \right] \right)
\end{align*}
% \begin{align}
%     {\bm h}_{\rm inst} &= f_{\rm FFN} ( {\bm h}_{\rm I}; {\bm h}_{\rm p} ), 
% \end{align}
where $f_{\rm FFN}$, $f_{\rm trm}$ and ${\bm h}_{\rm t}^{(n)}$ denote a feedforward network, $L_{\rm inst}$ transformer layers, and visual features extracted from ${\bm x}_{\rm t}^{(n)}$ (details are explained later), respectively.
\Update
Note that each  transformer layer consists of multi-head self attention layers \cite{vaswani2017attention} and feedforward networks (FFNs), and a FFN consists of a linear layer, normalization layer and activation function.
\Done

The CRFE handles multiple images of the surrounding context.
By handling context images that capture a broader range rather than just those that are in close proximity to the target bounding box, the proposed method can effectively model the relationship between the target object and its surrounding contextual environment.

% First, we obtain visual features ${\bm h}_{\rm t}^{(n)} \in {\mathbb R}^{768}$,  ${\bm h}_{\rm c}^{(n)} \in {\mathbb R}^{768}$, ${\bm h}_{\rm l}^{(n)} \in {\mathbb R}^{768}$ and ${\bm h}_{\rm r}^{(n)} \in {\mathbb R}^{768}$ from ${\bm x}_{\rm t}^{(n)}$, ${\bm x}_{\rm c}^{(n)}$, ${\bm x}_{\rm l}^{(n)}$ and ${\bm x}_{\rm r}^{(n)}$, respectively, using the pretrained CLIP image encoder (ViT/L-14).
% Then we obtain the embedding for image features ${\bm h}_{\rm targ}$ as follows:
% \begin{align}
%     {\bm h}_{\rm targ} &= f_{\rm img} ( {\bm h}_{\rm t}^{(n)}; {\bm h}_{\rm c}^{(n)}; {\bm h}_{\rm l}^{(n)}; {\bm h}_{\rm r}^{(n)} ),
% \end{align}
% where $f_{\rm img}$ denotes $L_{\rm img}$ transformer layers.
First, we obtain visual features ${\bm h}_{\rm t}^{(n)} \in {\mathbb R}^{768}$ and ${\bm h}_{\rm c}^{(n)} \in {\mathbb R}^{N_c \times 768}$ from ${\bm x}_{\rm t}^{(n)}$ and ${\bm x}_{\rm c}^{(n)}$, respectively, using the pretrained CLIP image encoder (ViT/L-14).
Then we obtain the embedding for image features ${\bm h}_{\rm targ}$ using $L_{\rm img}$ transformer layers.
Finally, we define the similarity between ${\bm x}_{\rm I}$ and $({\bm x}_{\rm t}^{(n)}, {\bm x}_{\rm c}^{(n)})$ using cosine similarity as follows:
\vspace{-1mm}
\begin{align*}
    s( {\bm x}_{\rm I}, {\bm x}_{\rm t}^{(n)}, {\bm x}_{\rm c}^{(n)} ) &= \frac{ {\bm h}_{\rm inst} \cdot {\bm h}_{\rm targ} }{ \| {\bm h}_{\rm inst} \|  \| {\bm h}_{\rm targ} \| }. 
\end{align*}
The output of our model is the ranked list of $T$ based on $s( {\bm x}_{\rm I}, {\bm x}_{\rm t}^{(n)}, {\bm x}_{\rm c}^{(n)} )$. 

The loss for each batch is defined as follows: 
\begin{align*}
    L_{\mathcal{B}} = - \frac{1}{|\mathcal{B}|} \sum_{{\bm x}_{\rm t}^{(n)} \in {\mathcal{B}}} \log{ \frac{ \exp{ (s( {\bm x}_{\rm I}, {\bm x}_{\rm t}^{(n)}, {\bm x}_{\rm c}^{(n)} ) )} }  { \sum_{j=1}^{|\mathcal{B}|}  \exp{(s( {\bm x}_{\rm I}, {\bm x}_{\rm t}^{(j)}, {\bm x}_{\rm c}^{(j)} ))} }  },
\end{align*}
where $|\mathcal{B}|$ denotes the batch size.
% Note that this loss function is equivalent to the scenario in which only ${\bm x}_{\rm I}$ is considered in infoNCE \cite{radford2021learning}.

%% file: section5.tex
\section{Dataset and Experimental Setup
\label{dataset}
}
\vspace{-1.5mm}

\subsection{Dataset}

\input{tab/results_val.tex}

\input{tab/results_test.tex}

To the best of our knowledge, no standard dataset exists for the LTRPO task.
For example, the REVERIE dataset \cite{qi2020reverie}, which is the standard dataset for object-goal navigation tasks, contains the important elements for the object localization task in that it deals with real indoor environmental images and object manipulation instructions.
However, the REVERIE dataset alone is unsuitable for the LTRTO task. Therefore, we built the Learning to Rank in Real Indoor Environments (LTRRIE) dataset, which is a new dataset for LTRPO tasks.

We built the LTRRIE dataset by the following steps: First, we collected instructions from the REVERIE dataset and panoramic images from the Matterport3D Simulator \cite{chang2017matterport3d}.
Then, we obtained the target bounding boxes by cropping the panoramic images  using the coordinates of the target objects provided in the REVERIE dataset. Note that we excluded samples with target bounding boxes at the edges of the former image because they may not have included the entire target object when cropped.
% The instructions in the REVERIE dataset were collected by over 1000 annotators using Amazon Mechanical Turk. The annotators were presented with animations of movement paths and randomly selected target objects. Then, they were instructed to create an instruction to manipulate the target object.

The LTRRIE dataset consists of 58 environments, 5501 instructions, and 4352 target bounding boxes. 
Note that the LTRRIE dataset consists of both complex (e.g., Fig. 1) and simple instructions (e.g., ``Bring me the white plant,'' ``Pick up the brown towel'').
It has a vocabulary size of 53118 words, a total of 103118 words, and an average sentence length of 18.78 words. 
The LTRRIE dataset includes 4210, 397 and 501 samples in the training, validation and test sets, respectively. 
Each set contains 50, 4 and 4 environments, respectively, with no duplication of environments.
Therefore, the samples included in the validation set can be regarded as unseen objects.
Compared to existing datasets (e.g., PFN-PIC \cite{hatori2018interactively}), the LTRRIE dataset is a comprehensive and diverse dataset that closely aligns with the real-world scenarios typically encountered by domestic service robots.
Note that we obtained the instructions for the training set from the REVERIE train set, and obtained the instructions for the validation and test sets  by dividing the val\_unseen set of the REVERIE dataset.
We used the training set to update the parameters of our model and the validation set to tune the hyperparameters. We evaluated our model on the test set.

\vspace{-1mm}
\subsection{Experimental Setup}

\vspace{-1mm}

We adopted the Adam optimizer ($\beta_1=0.9$,\, $\beta_2=0.999$) for training with learning rate of 2.0$\times 10^{-5}$ and batch size of 128.
In the transformer layer, we set $L_{\rm inst}=4, L_{\rm img}=4, \#A=4, \#H=768$, where $\#H$ and $\#A$ denote the hidden size and number of attention heads in the $f_{\rm inst}$ and $f_{\rm targ}$ transformer layers, respectively.

Our model had 47M trainable parameters and 24.49G multiply-add operations.
The proposed model was trained on an RTX 2080Ti with 11GB of GPU memory and an Intel Core i9 processor.
It took approximately an hour to train our model.
The inference time was approximately 20 ms/sample.
We evaluated our model on the validation sets every epoch. The final performance on the test set was achieved by the model with maximum recall@10 for the validation set.

%% file: tab/results_val.tex
\begin{table*}[htbp]
\centering
\normalsize
\vspace{1mm}
\caption{Quantitative comparison for the validation set. The best scores are in bold.}
\vspace{-2mm}
% \begin{tabular}{l|l|cc|ccccc}

% \hline
%     &                                  & \multicolumn{2}{c|}{Condition}                            & \multicolumn{5}{c}{Validation Set}                                                                            \\
%     &                                  & \multicolumn{1}{c}{w/ CNPE} & \multicolumn{1}{c|}{w/ ${\bm x}_{\rm c}^{(n)}$} & \multicolumn{1}{c}{MRR} & \multicolumn{1}{c}{R@1[\%]} & \multicolumn{1}{c}{R@5[\%]} & \multicolumn{1}{c}{R@10[\%]} & \multicolumn{1}{c}{R@20[\%]}  \\ \hline

\begin{tabular}{cc|cc|ccccc}
\hline
\multicolumn{2}{c|}{\multirow{2}{*}{Model}}                 & \multicolumn{2}{c|}{Condition}                            & \multicolumn{5}{c}{Validation Set}                                                                                                \\
\multicolumn{2}{l|}{}                                       & \multicolumn{1}{c}{w/ CNPE} & \multicolumn{1}{c|}{w/ ${\bm x}_{\rm c}^{(n)}$} & \multicolumn{1}{c}{MRR} & \multicolumn{1}{c}{R@1[\%]} & \multicolumn{1}{c}{R@5[\%]} & \multicolumn{1}{c}{R@10[\%]} & \multicolumn{1}{c}{R@20[\%]} \\ \hline

(a) & Baseline                         &                             & \checkmark                           & 0.433 $\pm$ 0.010           & 13.7 $\pm$ 0.6           & 47.1 $\pm$ 0.9           & 66.6 $\pm$ 1.4            & 82.0 $\pm$ 1.2          \\ \hline
(b) & \multirow{3}{*}{Proposed } & \checkmark                           &                             & 0.365 $\pm$ 0.022           & 11.5 $\pm$ 1.0           & 38.0 $\pm$ 3.2           & 54.9 $\pm$ 2.9            & 71.7 $\pm$ 1.4            \\
(c) &                                  &                             & \checkmark                           & 0.472 $\pm$ 0.007           & 16.8 $\pm$ 0.4           & \textbf{53.9 $\pm$ 1.2}           & \textbf{73.2 $\pm$ 0.1}            & \textbf{85.1 $\pm$ 0.7}             \\
(d) &                                  & \checkmark                           & \checkmark                           & \textbf{0.481 $\pm$ 0.014}           & \textbf{17.6 $\pm$  1.4}         & 50.8 $\pm$ 1.3           & 68.1 $\pm$ 1.3            & 83.0 $\pm$ 1.3           \\ \hline
\end{tabular}
\label{tab:results_val}
\vspace{-2mm}
\end{table*}

%% file: tab/results_test.tex
\begin{table*}[htbp]
\centering
\normalsize
\caption{Quantitative comparison for the test set. The best scores are in bold.}
\vspace{-2mm}
% \begin{tabular}{l|l|cc|ccccc}
% \hline
%     & & \multicolumn{2}{c|}{Condition}      & \multicolumn{5}{c}{Test Set}          \\
%     & & \multicolumn{1}{c}{w/ CNPE} & \multicolumn{1}{c|}{w/ ${\bm x}_{\rm c}^{(n)}$} & \multicolumn{1}{c}{MRR} & \multicolumn{1}{c}{R@1[\%]} & \multicolumn{1}{c}{R@5[\%]} & \multicolumn{1}{c}{R@10[\%]} & \multicolumn{1}{c}{R@20[\%]}  \\ \hline
\begin{tabular}{cc|cc|ccccc}
\hline
\multicolumn{2}{c|}{\multirow{2}{*}{Model}}                 & \multicolumn{2}{c|}{Condition}                            & \multicolumn{5}{c}{Test Set}                                                                                                \\
\multicolumn{2}{l|}{}                                       & \multicolumn{1}{c}{w/ CNPE} & \multicolumn{1}{c|}{w/ ${\bm x}_{\rm c}^{(n)}$} & \multicolumn{1}{c}{MRR} & \multicolumn{1}{c}{R@1[\%]} & \multicolumn{1}{c}{R@5[\%]} & \multicolumn{1}{c}{R@10[\%]} & \multicolumn{1}{c}{R@20[\%]} \\ \hline
(a) & Baseline   &       & \checkmark     & 0.415 $\pm$ 0.009& 14.0 $\pm$ 1.0& 45.3 $\pm$ 1.7& 63.8 $\pm$ 2.5 & 80.8 $\pm$ 2.0 \\ \hline
(b) & \multirow{3}{*}{Proposed } & \checkmark     &       &  0.373 $\pm$ 0.015& 12.1 $\pm$ 0.5& 39.6 $\pm$ 1.4& 56.1 $\pm$ 1.1 & 70.2 $\pm$ 0.7 \\
(c) & &       & \checkmark & 0.426 $\pm$ 0.004& 14.6 $\pm$ 0.4& 45.3 $\pm$ 0.5& 66.1 $\pm$ 1.7 & 80.6 $\pm$ 0.9 \\
(d) & & \checkmark     & \checkmark     & \textbf{0.501 $\pm$ 0.008}& \textbf{18.3 $\pm$ 1.0}          & \textbf{52.2 $\pm$ 1.4}& \textbf{69.8 $\pm$ 1.5} & \textbf{83.8 $\pm$ 0.6}\\ \hline
\end{tabular}
\label{tab:results_test}
\vspace{-5mm}
\end{table*}

%% file: section6.tex
\section{Experimental Results
\label{exp}
}
\vspace{-1mm}
\subsection{Results}
\vspace{-1mm}

To the best of our knowledge, limited studies exists in which the LTRPO task has been addressed. Thus, we developed a baseline method that extends CLIP \cite{radford2021learning} to handle the LTRPO task.
The baseline method takes ${\bm x}_{\rm I}$, ${\bm x}_{\rm t}$, and ${\bm x}_{\rm c'}$ as input. 
% Here, ${\bm x}_{\rm lcr}$ is obtained by concatenating ${\bm x}_{\rm c}$, ${\bm x}_{\rm l}$, and ${\bm x}_{\rm r}$ horizontally and resizing the result to 256$\times$256 pixels. 
Here, ${\bm x}_{\rm c'}$ is obtained by concatenating $N_c$ surrounding images of ${\bm x}_{\rm t}^{(n)}$ horizontally and resizing the result to 256$\times$256 pixels. 
${\bm x}_{\rm t}$ and ${\bm x}_{\rm c'}$ are input into the CLIP image encoder, and a simple MLP structure is used to obtain image features from these embeddings. Additionally, ${\bm x}_{\rm I}$ is input into the CLIP text encoder to obtain language features. Then we used these features to calculate the similarity using the same method explained in Section \ref{method}. The output is a ranked list of $T$, obtained based on this similarity.

Tables~\ref{tab:results_val} and \ref{tab:results_test} show the quantitative results of the baseline method, proposed method and ablation studies for the validation and test set, respectively. The values in the tables are the average and standard deviation over five trials. We explain the details of the ablation studies in Section \ref{sec:ablation}.
We evaluated the model using the MRR and recall@k (k $=1,5,10,20$). We used the MRR as the primary metric. 
% We evaluated the model by mean reciprocal rank (MRR) and recall@k. We used the MRR as the primary metric. å
MRR is defined as follows:
\vspace{-2mm}
\begin{align*}
    \rm{MRR} &= \frac{1}{N_{\rm inst}} \sum_{j=1}^{N_{\rm inst}} \frac{1}{r_1^{(j)}}\,, 
\end{align*}
where $N_{\rm inst}$ and ${r_1^{(j)}}$ denote the number of instructions and the rank of the first relevant sample in the list of retrieved samples, respectively. 
Recall@k is defined as follows: 
\vspace{-2mm}
\begin{align*}
    {\rm Recall@k} &= \frac{1}{N_{\rm inst}} \sum_{j=1}^{N_{\rm inst}} \frac{|A_j \cap B_j |}{|A_j |} \,, 
\end{align*}
where $A_j$ and $B_j$ denote the set of relevant samples to be searched and the set of the top--$k$ retrieved samples, respectively.
We used the MRR and recall@k because they are standard metrics in learning--to--rank settings \cite{liu2009learning}.

\input{fig/qualitative_success.tex}

\input{fig/qualitative_failure.tex}

Table~\ref{tab:results_test} shows that the MRR for the baseline method (a) and the proposed method (d) were 0.415 and 0.501, respectively. Therefore, the proposed method (d) outperformed the baseline method (a) on the test set by 0.086 points in terms of the MRR. 
Similarly, the proposed method also outperformed the baseline method in terms of recall@k (k $\leq 20$).
For all the above results except for recall@20, there were statistically significant differences in performance ($p<0.05$) between our method and the baseline method.

% \subsection{Qualitative Results}

Fig.~\ref{fig:success} shows the qualitative results. 
Fig.~\ref{fig:success} (a) presents the top--3 samples in the output ranked list for ${\bm x}_{\rm I}$ ``Go to the bathroom with a picture of a wagon and bring me the towel directly across from the sink.'' 
For this sample, the MRR was 1 and 0.1 for the proposed method and the baseline method, respectively.
% It is worth noting that the ``sink'', which was mentioned in ${\bm x}_{\rm I}$, was not present in ${\bm x}_{\rm c}$, whereas it was present in ${\bm x}_{\rm r}$. 
It is worth noting that the ``sink'', which was mentioned in ${\bm x}_{\rm I}$, was only presented in edge of ${\bm x}_{\rm c}^{(n)}$. 
Therefore, this result suggests that the introduction of the CRFE, which handles context images that capture a broader range, was beneficial to  performance.

Fig.~\ref{fig:success}(b) shows the top--3 samples in the output ranked list for ${\bm x}_{\rm I}$ ``Go to the hallway on level 1 that is lined with wine bottles and pull out the high chair closest to the wine bottles at the second table from the door.'' For this sample, MRR was 1 and 0.091 for the proposed method and baseline method, respectively.
% In this sample, ${\bm x}_{\rm I}$ includes a complex referring expression ``the high chair closest to the wine bottles at the second table from the door.''
% However, the proposed model successfully retrieved the appropriate ${\bm x}_{\rm t}$ as ``rank 1.'' 
% The proposed model successfully retrieved the appropriate ${\bm x}_{\rm t}$ as ``rank 1'' for the complex referring expression: ``the high chair closest to the wine bottles at the second table from the door.''
The proposed model successfully retrieved the appropriate ${\bm x}_{\rm t}$ as top--1 result for the complex referring expression: ``the high chair closest to the wine bottles at the second table from the door.''
Therefore, this result suggests that the introduction of the CNPE module to model the relationship between noun phrases and target objects was beneficial to performance.

Fig.~\ref{fig:failure} illustrates a sample for which the model failed to retrieve the desired image as specified in ${\bm x}_{\rm I}$, ``Proceed to the hallway on level 2 with the basketball and level painting above the open book.'' The instruction required the model to retrieve an image of the ``painting above the open book,'' but the model was unsuccessful in retrieving images of paintings that were not positioned near an open book. We hypothesize that this failure was because of the lack of a ``book'' in ${\bm x}_{\rm c}^{(n)}$, which made it challenging for the model to accurately search using a referring expression.

\vspace{-2mm}
\subsection{Ablation Study \label{sec:ablation}}
\vspace{-1mm}

We set the following two conditions for the ablation studies to demonstrate the effectiveness of our proposed method. Tables~\ref{tab:results_val} and \ref{tab:results_test} present the quantitative results for the ablation studies. 

{ 
\setlength{\leftmargini}{0pt}
\begin{itemize}
\item[] \textbf{${\bm x}_{\rm c}^{(n)}$ ablation: } We removed ${\bm x}_{\rm c}^{(n)}$ to investigate the performance when only ${\bm x}=({\bm x}_{\rm I}, T)$ was used as input. 
\Update
Table~\ref{tab:results_test} \Done shows that the MRR decreased by 0.128 points under condition (b) on the test set. This result indicates that introducing CRFE to handle ${\bm x}_{\rm c}^{(n)}$ was most beneficial to performance. We believe that the introduction of CRFE was effective due to its capability to handle contextual images that capture a broader range, enabling the transformer layer subsequent to CRFE to effectively model the relationship between the target object and its surrounding  environment.
\item[] \textbf{CNPE ablation: } We removed the CNPE to investigate the influence on performance. \Update Table~\ref{tab:results_test} \Done shows that the MRR decreased by 0.075 points under condition (c) on the test set. This results indicates that the introduction of the CNPE, to handle noun phrases extracted from complex instructions, was also beneficial to performance. We believe that the introduction of CNPE proved effective because it enabled the model to effectively model the relationship between the image of the target object and the phrases associated with the object within the environment.
\end{itemize}
}

\input{tab/error_analysis.tex}

\vspace{-4mm}
\subsection{Error Analysis}
\vspace{-1mm}
% We defined samples with MRR less than 0.1 as failure cases.
We analyzed samples with an MRR less than 0.1, which we defined as failed cases.
We identified 171 failed cases, and conducted a detailed analysis of the top--10 retrieved results for 20 of these failed cases.
Table~\ref{tab:erroranalysis} categorizes the failed cases.
We classified the causes of failure into the following six groups:
(i) REC Error (RECE): This refers to the case in which the top retrieved results did not comply with the referring expression. 
(ii) Target Noun Phrase Selection Error: 
This refers to the case in which the model did not retrieve the target object but retrieved one of the landmark objects referred to in the sentence.
(iii) Object Grounding Error: This refers to the case in which the model has achieved inadequate performance for object grounding. 
(iv) Annotation Error: This refers to the case in which the annotation for the target bounding box or the instruction was incorrect.
(v) Ambiguous Instruction: This refers to the case in which the ${\bm x}_{\rm I}$ was ambiguous and the top search results could not be definitively deemed incorrect.
(vi) Others: This refers to errors that we  could not classify under the previously mentioned categories. 
Table~\ref{tab:erroranalysis} shows that the main bottleneck was RECE.
As a possible solution, we could effectively address this error by introducing a module that handles map information.

%% file: fig/qualitative_success.tex
% \begin{figure*}[t]
%     \centering
%     % \includegraphics[width=152mm]{fig/fig6-1.jpg}
%     \includegraphics[width=\linewidth]{fig/fig6-1-1.jpg}
%     \vspace{-2mm}
%     \caption{Qualitative Results. The topmost images show the ground truth target image and the others show the top three ranked images. The yellow-bordered area and the green dotted-line area represent ${\bm x}_{\rm t}$ and contextual objects other than ${\bm x}_{\rm t}$, respectively. The lower left and lower right images within the black borders represent ${\bm x}_{\rm l}$ and ${\bm x}_{\rm r}$, respectively. Note that some of ${\bm x}_{\rm l}$ and ${\bm x}_{\rm r}$ are omitted for visibility.}
%     \label{fig:success}
% \end{figure*}

\begin{figure*}[htbp]
    \centering
    \vspace{2mm}
    \includegraphics[width=150mm]{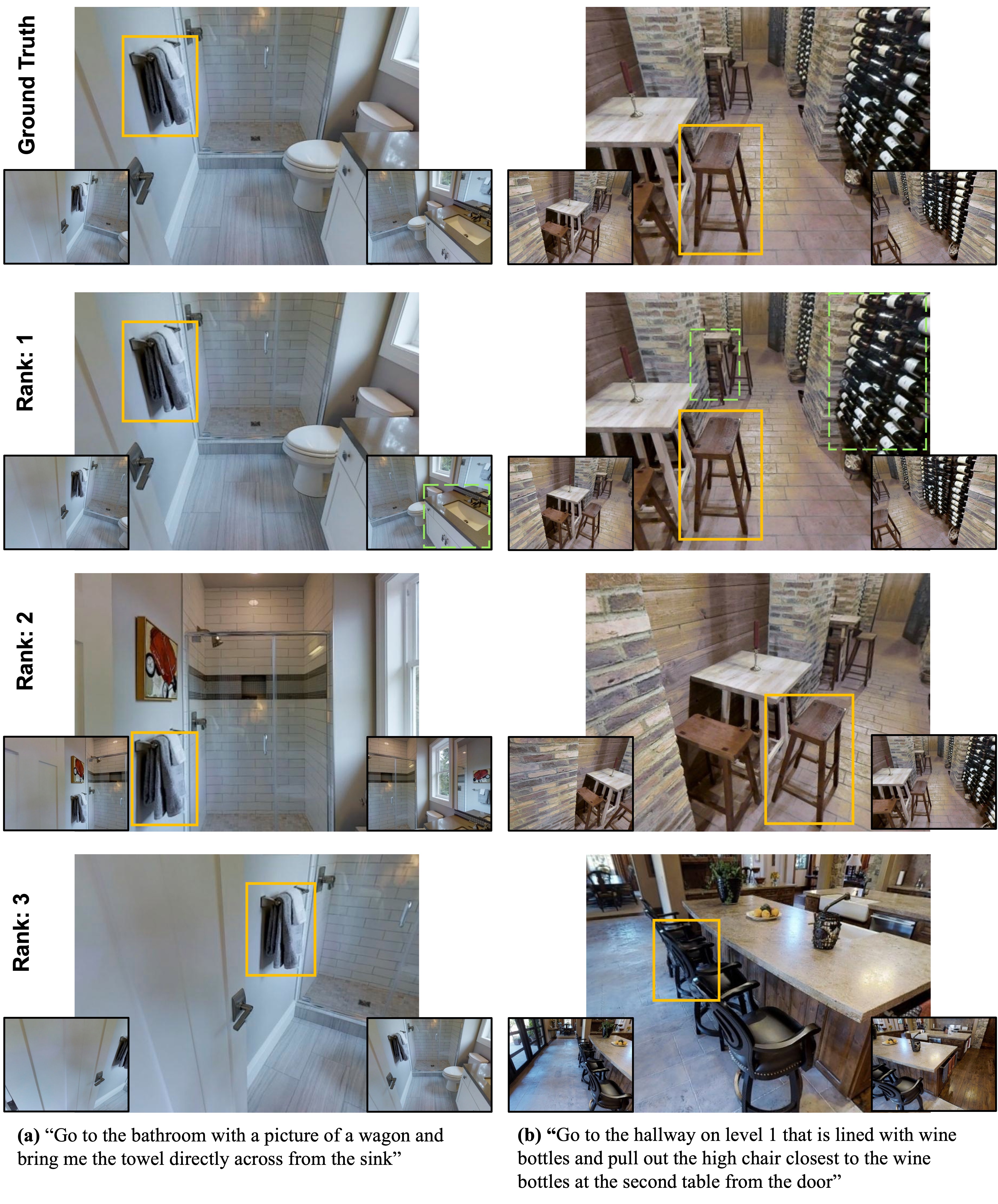}
    \vspace{-2mm}
    \caption{Qualitative Results. The topmost images show the ground truth target image and the others show the top--3 ranked images. The yellow-bordered area and the green dotted-line area represent ${\bm x}_{\rm t}$ and contextual objects other than ${\bm x}_{\rm t}$, respectively. }
    \vspace{-10mm}
    \label{fig:success}
\end{figure*}

% The lower left and lower right images within the black borders represent ${\bm x}_{\rm l}$ and ${\bm x}_{\rm r}$, respectively.

%% file: fig/qualitative_failure.tex
\begin{figure*}[htbp]
\vspace{-3mm}   
    \centering
    \includegraphics[width=155mm]{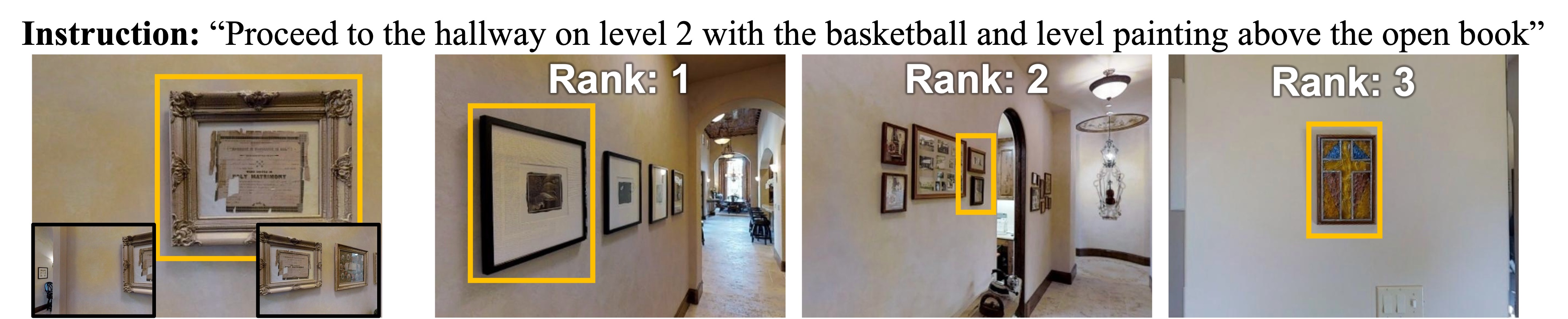}
    \caption{A failed sample. The leftmost image shows the ground truth target image and the others show the top--3 ranked images. The model was unsuccessful in retrieving images of painting that were positioned near an open book.}
    \label{fig:failure}
    \vspace{-5mm}
\end{figure*}

%% file: tab/error_analysis.tex
\begin{table}[t]
\centering
\normalsize
\caption{Results for error analysis.}
\vspace{-2mm}
\begin{tabular}{lc}
\hline
Error & \#Samples \\ \hline
Referring Expression Comprehension Error & 6 \\
Target Noun Phrase Selection Error & 5 \\
Object Grounding Error & 3 \\
Annotation Error & 2 \\
Ambiguous Instruction & 2 \\
Others & 2 \\\hline
\end{tabular}
\label{tab:erroranalysis}
\vspace{-1mm}
\end{table}

% The main bottleneck was referring expression comprehension error.

%% file: section6-2.tex
\vspace{-2mm}
\section{Physical Experiments}
\vspace{-1mm}

\input{fig/env_and_ycb}

% \subsection{DSR, Environment and Procedure}
We used the Human Support Robot (HSR) \cite{yamamoto2019hsr}, developed by the Toyota Motor Corporation (shown in Fig.~\ref{fig:env_and_ycb}(a)). The physical experiments involved YCB objects \cite{calli2015ycb} (Fig.~\ref{fig:env_and_ycb}(b)).
% We used the 20 objects in the upper group as target objects.
Fig.~\ref{fig:env_and_ycb}(a) shows the environment used in the physical experiments, which was based on the standardized environment of the World Robot Summit 2020 Partner Robot Challenge/Real Space  \cite{wrs2020x}, an international robotics competition that evaluated tidying tasks in domestic settings. The environment's dimensions measured $6.0 \times 4.0 {\rm m}^2$, comprising six specified types of furniture pieces arranged as depicted in Fig.~\ref{fig:env_and_ycb}(a). 
% (WRS2020RS)

% \vspace{-2mm}
% \subsection{Procedure}
% In the following section, experimental settings about the physical experiments are explained. 
The procedure for the physical experiments are as follows:
Firstly, we generated 5 distinct object placement patterns, ensuring all objects were placed on furniture pieces. Subsequently, we provided the DSR with a total of 50 English instructions.
% , such as ``Pick up the apple next to the green cup and bring it to me.''
The DSR's behavior was designed as follows: It began by collecting environment images while navigating through waypoints from the initial position, employing standard path planning and navigation methods based on a pre-provided map. Each waypoint allowed the DSR to face each furniture piece and capture the target object from multiple viewpoints using an Asus Xtion Pro camera mounted on its head. We performed zero-shot transfer using our model trained on the LTRRIE dataset.
Note that this indicates that the results of the physical experiments can be viewed as demonstrating the generalization capability of the proposed model.
For grasping actions, the grasping point was determined based on the depth image and the bounding box of the target object. Using the intrinsic camera parameters, a point cloud was obtained from the bounding box of the depth image and transformed into the camera coordinate system. The median position in each coordinate axis was then identified as the grasping point.
\Update
It took approximately 4 seconds to execute the process in real-time, including preprocessing, data transfer and inference, with our extended version.
\Done
% \vspace{-2mm}
% \subsection{Results}

Table \ref{tab:results_real} shows the quantitative results of the physical experiments. We used MRR, recall@10 and task success rate as the evaluation metrics. 
To enable the user/operator to easily select the target object in standard UI setting, we only considered ranks within the top--10 for calculating the MRR.
% Considering the importance of presenting the correct target object to the user within the top 10 results, we only considered ranks within the top 10 for calculating the MRR.
Table \ref{tab:results_real} shows that the MRR and recall@10 were 0.45 and 88, respectively. 
% Therefore, the model was able to rank the appropriate target object in top 10 results 88\% of the time. 
Also, the task success rate for object retrieval and manipulation was 80\%.
Fig. \ref{fig:qualitative_real} shows qualitative results of the physical experiment, where the red rectangles represent candidate object regions. The top--3 samples in the ranked list of outputs are displayed, along with surrounding images for each candidate object area. 
% In Fig. \ref{fig:qualitative_real}(a), the target object, a spam can, was appropriately ranked as ``rank 1,'' and the user selected the first-ranked object region from the top 10. 
In Fig. \ref{fig:qualitative_real}(a), the target object, a spam can, was appropriately retrieved as the top--1 result, and the user selected the appropriate object region.
The robot successfully grasped the spam can and delivered it to the user. 
% Similarly, in Fig. \ref{fig:qualitative_real}(b), the target object, a yellow cup, was precisely ranked as ``rank 1,'' and the yellow cup selected by the user was successfully delivered.
Similarly, in Fig. \ref{fig:qualitative_real}(b), the target object, a yellow cup, was appropriately retrieved as the top--1 result, and the yellow cup selected by the user was successfully delivered.

\input{tab/result_real}

\input{fig/qualitative_real}

%% file: fig/env_and_ycb.tex
\begin{figure}[t]
    \centering
    \includegraphics[width=80mm]{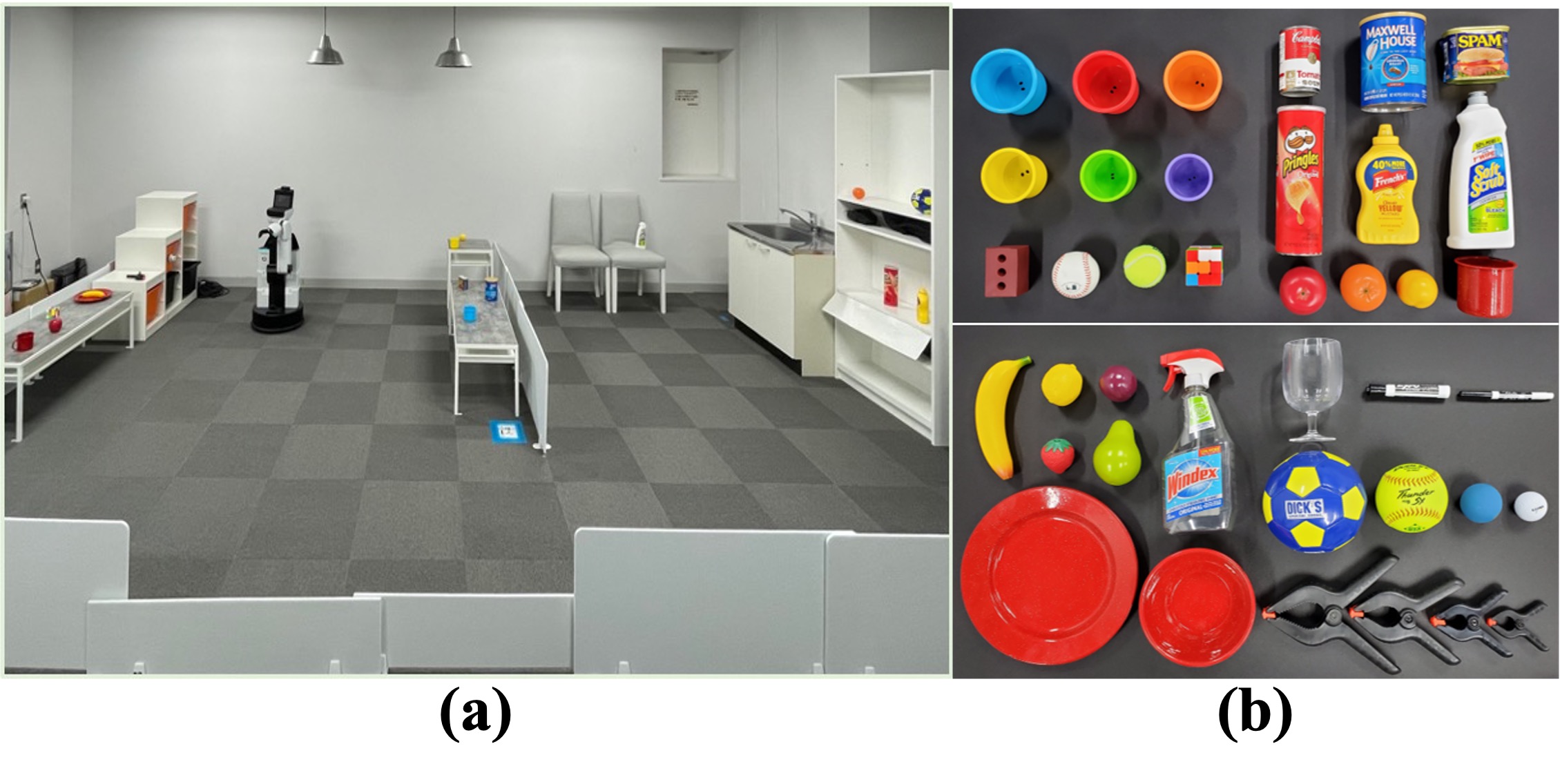}
    \vspace{-3mm}
    \caption{(a) DSR platform in the experiment environment and (b) YCB objects used in  the physical experiments.}
    \label{fig:env_and_ycb}
    \vspace{-5mm}
\end{figure}

%% file: tab/result_real.tex
\begin{table}[t]
\centering
\normalsize
\caption{Task success rates in the physical experiments. }
\vspace{-1mm}
\begin{tabular}{cccc}
\hline
MRR & Recall@10 {[}\%{]}       & SR {[}\%{]}           \\ \hline
0.45  & 88 &  80 (40 / 50)  \\ \hline
&  & & \multicolumn{1}{l}{}                  
\end{tabular}
\vspace{-9mm}
\label{tab:results_real}
\end{table}

% $N_{\rm success}$ and $N_{\rm attempts}$ denote the number of attempts and successes, respectively.

%% file: fig/qualitative_real.tex
% \begin{figure*}[t]
%     \centering
%     \includegraphics[width=140mm]{fig/fig6-5.jpg}
%     \caption{Qualitative results of the physical experiments. The leftmost image shows the ground truth target image and the others show the top--3 ranked images. For (a) and (b), the DSR successfully retrieved images of target objects.}
%     \label{fig:qualitative_real}
%     \vspace{-6mm}
% \end{figure*}

\begin{figure}[t]
    \centering
    \includegraphics[width=85mm]{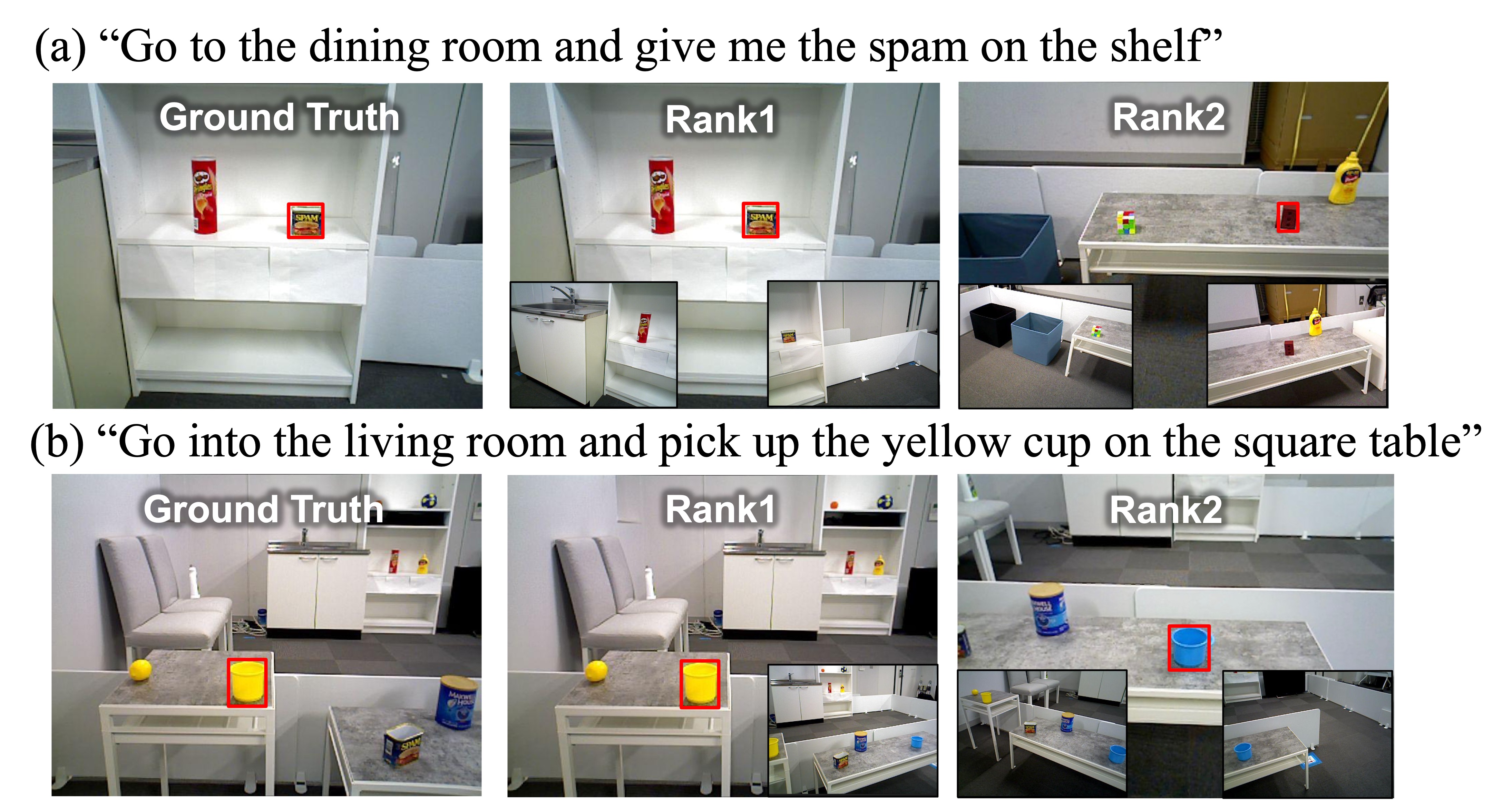}
    \vspace{-0.5mm}
    \caption{Qualitative results of the physical experiments. The leftmost image shows the ground truth target image and the others show the top--2 ranked images. For (a) and (b), the DSR successfully retrieved images of target objects.}
    \label{fig:qualitative_real}
    \vspace{-5mm}
\end{figure}

%% file: section7.tex
\vspace{-1mm}
\section{Conclusions}
\vspace{-1mm}

In this paper we focused on the LTRPO task, which is a task to retrieve images of a target object taken by DSRs according to given instructions in a human-in-the-loop setting. Our method outperformed the baseline method for all  metrics on the LTRRIE  dataset. Also in physical experiments, our method achieved task success rate of 80\%, which indicates that language comprehension and action execution can be integrated in a physical robot.
% We emphasize the following contributions of this study:
% \begin{itemize}
%     \item We proposed MultiRankIt, which is a novel approach for identifying target objects from open-vocabulary user instructions in a human-in-the-loop setting.
%     \item We introduced the Crossmodal Noun Phrase Encoder to model the relationship between phrases that contain referring expressions and target bounding box.
%     \item We introduced the Crossmodal Region Feature Encoder to model the relationship between the target object and multiple images of its surrounding contextual environment.
%     \item Our method outperformed the baseline method for all  metrics on the LTRRIE  dataset.
%     \item  In physical experiments, our method achieved task success rate of 80\%, which indicates that language comprehension and action execution can be integrated in a physical robot. 
% \end{itemize}
In future work, we plan to extend our model by introducing a new module to handle map information.

% Most data-driven approaches for multimodal language understanding require large-scale datasets.  
% However, building such a dataset is time-consuming and costly.
% In this study, we proposed the ABEN, which generates fetching instructions from images. Target use cases include generating and augmenting datasets of image--sentence pairs.

% The following contributions of this study can be emphasized:
% \begin{itemize}
%  \item The ABEN extends the  Multi-ABN by introducing a linguistic branch and a generation branch,  to model the relationship between subwords. 
%  \item  The ABEN combines attention branches and BERT-based subword embedding for sentence generation. 
% \end{itemize}

% Future studies will investigate the application of the ABEN to real-world settings.